\documentclass[conference]{IEEEtran}
\IEEEoverridecommandlockouts
\usepackage{cite}
\usepackage{url}

\usepackage{hyperref} %
\hypersetup{
    colorlinks=false,
    pdfborder={0 0 0},
}
\usepackage{amsmath,amssymb,amsfonts}
\usepackage{algorithmic}
\usepackage{graphicx}
\usepackage{textcomp}
\usepackage{xcolor}
\def\BibTeX{{\rm B\kern-.05em{\sc i\kern-.025em b}\kern-.08em
    T\kern-.1667em\lower.7ex\hbox{E}\kern-.125emX}}
\begin{document}

\title{Scaling Equilibrium Propagation to Deeper Neural Network Architectures\\
\thanks{This work was supported in part by the RISC-V Knowledge Centre of Excellence (RKCoE), sponsored by the Ministry of Electronics and Information Technology (MeitY).}
}

\author{\IEEEauthorblockN{Sankar Vinayak E P}
\IEEEauthorblockA{\textit{Dept. of Computer Science and Engineering,} \\
\textit{Indian Institute of Technology, Madras}\\
\href{mailto:cs24m041@smail.iitm.ac.in}{cs24m041@smail.iitm.ac.in}}
\and
\IEEEauthorblockN{Gopalakrishnan Srinivasan }
\IEEEauthorblockA{\textit{Dept. of Computer Science and Engineering,} \\\textit{Robert Bosch Centre for Data Science and AI,} \\
\textit{Indian Institute of Technology, Madras}\\
\href{mailto:sgopal@cse.iitm.ac.in}{sgopal@cse.iitm.ac.in}}
}

\maketitle

\begin{abstract}
Equilibrium propagation has been proposed as a biologically plausible alternative to the backpropagation algorithm. The local nature of gradient computations, combined with the use of convergent RNNs to reach equilibrium states, make this approach well-suited for implementation on neuromorphic hardware. However, previous studies on equilibrium propagation have been restricted to networks containing only dense layers or relatively small architectures with a few convolutional layers followed by a final dense layer. These networks have a significant gap in accuracy compared to similarly sized feedforward networks trained with backpropagation. In this work, we introduce the Hopfield-Resnet architecture, which incorporates residual (or skip) connections in Hopfield networks with clipped $\mathrm{ReLU}$ as the activation function. The proposed architectural enhancements enable the training of networks with nearly twice the number of layers reported in prior works. For example, Hopfield-Resnet13 achieves 93.92\% accuracy on CIFAR-10, which is $\approx$3.5\% higher than the previous best result and comparable to that provided by Resnet13 trained using backpropagation. Our implementation source code is available at \href{https://github.com/BrainSeek-Lab/Scaling-Equilibrium-Propagation-to-Deeper-Neural-Network-Architectures}{https://github.com/BrainSeek-Lab/Scaling-Equilibrium-Propagation-to-Deeper-Neural-Network-Architectures}.

\end{abstract}

\section{Introduction}
Advancements in deep learning, powered by artificial neural networks (ANNs), have formed the backbone of rapid growth in modern artificial intelligence. These improvements can be attributed to backpropagation (BP) based training algorithms, which solve the error credit assignment problem by applying chain rule to propagate gradients from the output layer. However, backpropagation is regarded as biologically implausible since the underlying gradient computations are non-local and require access to global network information. Alternative algorithms such as forward propagation~\cite{forwardforward}, predictive coding~\cite{PCN}, and no-prop~\cite{noprop} have been proposed to address the non-locality issue of backpropagation. These emerging algorithms offer the potential for on-device learning with higher energy efficiency compared to backpropagation.

Contrastive Hebbian~\cite{contrastive_hebbian} learning is one such algorithm that has been shown to converge to a steady state over time for static input, effectively operating as a static convergent recurrent neural network (RNN). Equilibrium propagation (EP)~\cite{BengioEP} is a category of contrastive Hebbian learning rule based on the network energy function, which is computed using both the neuronal states and the network parameters. The EP algorithm operates in two phases. Initially, the network dynamics are governed exclusively by the energy function. Subsequently, a weak clamping force proportional to the loss function of the output layer is applied, via a weighting parameter, causing the network to reach an equilibrium state with reduced loss and lower total energy. The combination of input layer clamping and weak output layer clamping enables gradient computation without requiring backpropagation across the entire network. The credit assignment problem is solved by leveraging the difference in the gradient of the energy function with respect to network parameters between the two phases, thereby providing a learning rule with greater biological plausibility. Previous works have implemented EP in hardware, demonstrating its potential for on-device learning~\cite {crossbar, onchip}.

The primary limitation of EP is that, although the gradient computed by this method closely matches that obtained using backpropagation through time~\cite{matchBPTT}, its performance is typically lower than that of a comparable feedforward network trained with backpropagation. In particular, the bias on the estimated gradient arising from the theoretical requirement for infinitely small nudging introduces noise into the computations, corrupting the gradient and thereby hindering learning. This issue was addressed by methods such as centered equilibrium propagation (CEP)~\cite{ScalingToConv} and holomorphic equilibrium propagation (HEP)~\cite{HEP}, which increase the number of nudging phases during network training. However, these methods have been validated primarily on shallow networks ($\leq$ 6 trainable layers), which still exhibit performance degradation compared to that achieved through backpropagation. We propose architectural enhancements to improve the scalability and performance of EP-based training. Overall, the key contributions of our work are as follows.
\begin{itemize}
    \item We propose \textit{clipped ReLU} activation function to simplify the energy function and gradient computation.
    \item We introduce the residual Hopfield network architecture, termed \textit{Hopfield-Resnet}, which consists of residual or skip connections that enable EP-based methods to successfully train deeper networks ($>$ 12 layers) with minimal performance loss relative to backpropagation baselines.
    \item We validate the proposed Hopfield-Resnet architecture and training methodology across the CIFAR-10, CIFAR-100, and Fashion MNIST datasets.
\end{itemize}

\section{Related Works}
\subsection{Static Convergent RNN}
In a supervised training setting, given an input ($x$), the network is trained to produce the required output ($y$). Equilibrium propagation, on the contrary, uses a convergent RNN, wherein the network evolves its neuronal state $s$ to a steady state $s_*$ over time for the input $x$. An energy function $\Phi$ is computed based on the neuronal state $s_t$ and the network parameters $\theta$. The network evolves as
\begin{equation}
    s_{t+1}=\frac{\partial \Phi(x,s_t,\theta)}{\partial s},
\end{equation}
and the equilibrium state $s_*(=s_{t+1}=s_t)$ is specified by
\begin{equation}
    s_*=\frac{\partial \Phi(x,s_*,\theta)}{\partial s}.
\end{equation}
That is, the network converges to steady state. The parameters are updated in such a way that, at equilibrium, the state of the output neurons matches the expected output $y$.

\subsection{Equilibrium Propagation}
Equilibrium propagation (EP)~\cite{BengioEP} was originally proposed for continuous-time dynamics. Subsequent works extended the method to support discrete-time dynamics~\cite{matchBPTT}. EP computes the gradient in two phases. During the first phase (\textit{free} phase), the input layers of the network are clamped to $x$, and the network is allowed to evolve based only on the energy function without any label information. In the second phase (or \textit{weakly clamped} phase), the output layer is perturbed by an additional term proportional to gradient of the loss $L$ with respect to the neuronal states, with its magnitude scaled by the parameter $\beta$. The updated neuronal dynamics is specified by
\begin{equation}
    s_{t+1}=\frac{\partial \Phi(x,s_t,\theta)}{\partial s}+\beta\frac{\partial L(x,s_t,\theta)}{\partial s}.
\end{equation}
The network then converges to new steady state $s_*^\beta$. It has been shown that if the energy function is differentiable with respect to both $\beta$ and the network parameters $\theta$, the gradient of the loss function $L$ with respect to the parameters can be obtained from the gradient of the energy function at equilibrium as

\begin{equation}
\label{eq:loss_wrt_parameters}
-\frac{\partial L}{\partial \theta}
= \frac{1}{\beta} \left[ 
\frac{\partial \Phi(x, s_*^\beta, \theta)}{\partial \theta}
- \frac{\partial \Phi(x, s_*, \theta)}{\partial \theta}
\right].
\end{equation}
Equation~\ref{eq:loss_wrt_parameters} holds as $\beta\rightarrow 0$.

\subsection{Centered Equilibrium Propagation}
\label{sec:CEP}
The theoretical requirement for an infinitesimally small $\beta$ (refer Equation~\ref{eq:loss_wrt_parameters}) introduces gradient estimator bias for non-zero values of beta, thereby limiting the practical application of the vanilla EP method. The centered equilibrium propagation (CEP) algorithm~\cite{ScalingToConv} mitigates this by computing the gradient using both $+\beta$ and $-\beta$, and using a second-order approximation for improving the gradient estimation. The equation then becomes 
\begin{equation}
-\frac{\partial L}{\partial \theta}
= \frac{1}{2\beta} \left[
\frac{\partial \Phi(x, s_*^{+\beta}, \theta)}{\partial \theta}
- \frac{\partial \Phi(x, s_*^{-\beta}, \theta)}{\partial \theta}
\right].
\end{equation}
Further research has focused on computing stable equilibrium at multiple points within a finite-sized oscillation of radius $\beta$, along the complex plane, to further reduce the bias in gradient estimation~\cite{HEP}. This approach leads to the equation
\begin{equation}
\frac{\partial L}{\partial \theta}
= \frac{1}{T|\beta|} \int_{0}^T
\frac{\partial \Phi}{\partial \theta}
\left( \theta, s_*^{\beta(t)}, \beta(t) \right)
e^{-2 i \pi t / T} \, dt,
\end{equation}
where $T$ is the period of the teaching signal (or the number of points for computing the equilibrium), $t\in[0,T]$, and $\beta(t)=|\beta|e^{-2 i \pi t / T}$.

\subsection{Convolutional Network with Equilibrium Propagation}
The theory of EP, originally proposed for dense networks, has been extended to networks with convolutional operations. In a network with dense and convolutional layers, the energy function $\Phi$ is the sum of the contributions from each part~\cite{ScalingToConv}, given by
\begin{equation}
\label{equation:energyfn}
\begin{split}
\Phi(\theta,\{s^n\})
&= \sum_{n=0}^{N_{\mathrm{conv}}-1}
s^{n+1} \cdot \mathcal{P}\!\left( w_{n+1}\star s^n \right) \\
&\quad + \sum_{n = N_{\mathrm{conv}}}^{N_{\mathrm{tot}}-1}
s^{n+1\top} w_{n+1} s^n
\end{split}
\end{equation}
where $s^n$ is the neuronal state per layer, $w$ denotes the weight matrix, $\mathcal{P}$ represents the pooling operation, and $\star$ indicates the convolution operation. The layer-wise state evolution can then be formulated as
\begin{align}
s^{n}_{t+1} &= \sigma\Big( 
    \mathcal{P}\big( w_n \star s^{n-1}_{t} \big) 
    + \tilde{w}_{n+1} \star \mathcal{P}^{-1}\big( s^{n+1}_{t} \big) 
\Big), \\
&\quad \text{for } 1 \leq n \leq N^{\mathrm{conv}} \nonumber \\[1ex]
s^{n}_{t+1} &= \sigma\Big( 
    w_n  s^{n-1}_{t} 
    + w_{n+1}^\top  s^{n+1}_{t} 
\Big), \\
&\quad \text{for } N^{\mathrm{conv}} < n < N^{\mathrm{tot}} \nonumber
\end{align}
where  $\sigma$ is the activation function used, $\tilde{w}$ is the flipped kernel used for transpose convolution, and $\mathcal{P}^{-1}$ is the inverse pooling operation.

\subsection{Asynchronous Update of Neuronal States}
The inherent characteristics of state update dynamics in EP method cause extended convergence times for larger network architectures. Gradient computation without complete steady-state convergence introduces significant noise into the learning process. One method used to reduce the time to equilibrium is the asynchronous update of the neuronal state~\cite{EnergyCompare}. Instead of performing global state updates following a single energy evaluation, this method uses an alternating update scheme in which energy is computed to first update even-indexed layers, followed by energy recomputation with the new states to adjust odd-indexed layers within each iteration cycle. Although there is no proof of convergence for this technique, it helps reduce the number of discrete iterations needed to reach equilibrium, thus shortening the overall training time.

\section{Proposed Architecture}
Previous works utilizing equilibrium propagation conducted experiments using relatively smaller networks, such as $VGG5$ with $4$ convolutional layers and $1$ dense layer~\cite{vgg, EnergyCompare, ScalingToConv}, or $VGG6$ with $2$ dense layers~\cite{HEP}, or networks consisting only of dense layers~\cite{BengioEP}, limiting the achievable depth and constraining performance across benchmark datasets. Our work addresses the scalability bottleneck by introducing \textit{Hopfield-Resnet}, featuring residual connections between layers and modifying the activation function, both of which contribute towards improved training of deeper networks with higher accuracy.

\subsection{Scaling Deeper with Residual Hopfield Network}
\begin{figure}
    \centering
    \includegraphics[width=0.7\linewidth]{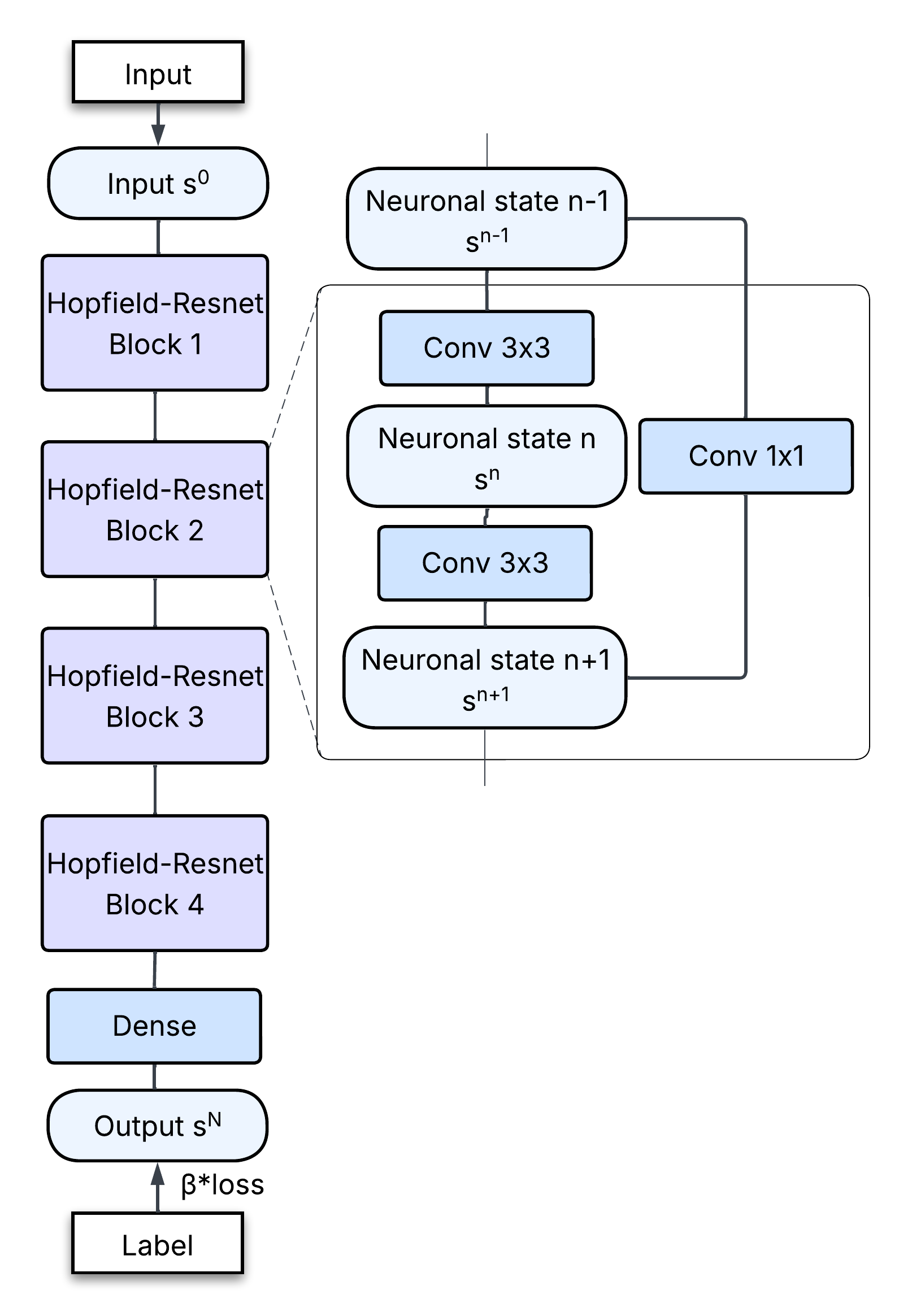}
    \caption{Architecture of Hopfield-Resnet13, consisting of four Hopfield-Resnet blocks, each containing three convolutional layers (two of them in the main pathway and one skip connection), followed by a dense output layer.}
    \label{fig:hopfieldresnet}
\end{figure}

The non-feedforward architecture of convergent RNNs imposes a primary scalability limitation. As the network becomes deeper, it requires more parameters and a longer time to reach steady state, thereby making training increasingly challenging. Our experiments indicate that, beyond a certain depth, adding more layers yields diminishing returns. Residual connections are a widely used technique for scaling networks deeper while ensuring training convergence~\cite{resnet}. In the case of a Hopfield network~\cite{hopfieldnetwork}, there exists indirect interaction between different layers through the energy function. We further propose a residual Hopfield network, referred to as \textit{Hopfield-Resnet}, which incorporates residual connections to improve the scalability, as shown in Fig.~\ref{fig:hopfieldresnet}. The basic Hopfield-Resnet block consists of three convolutional operations: two forming the main pathway and one skip connection that directly links the final state of the preceding block to the final state of the current block. %
The convolution on the main path uses $3\times3$ kernels, while the one in skip connection uses $1\times1$ kernels. Residual connections are implemented in two ways: direct identity connections and connections using a $1\times1$ kernel. Both approaches were tested, with the latter demonstrating better performance and therefore selected for the experiments. The network architecture used in this work consists of four Hopfield-Resnet blocks and a dense layer. This design yields $13$ sets of trainable parameters, with $12$ convolution layers, dense output layer, and $9$ neuronal states which can be updated.%

The Hopfield-Resnet architecture enhances interactions between neuronal states during energy computation while preserving the energy function previously defined in Equation~\ref{equation:energyfn}. The neuronal state update equation is modified so that, instead of considering only adjacent states, it accounts for all paths and states that directly interact with the current state. This leads to summation over all such interaction paths, as described by
\begin{align}
s^{n}_{t+1} &= \sigma\Big( \sum_{i=pre(n)}
    \mathcal{P}\big( w_i \star s^{i}_{t} \big) 
    +\sum_{j=post(n)} \tilde{w}_{j} \star \mathcal{P}^{-1}\big( s^{j}_{t} \big) 
\Big), \\
&\quad \text{for } 1 \leq n \leq N^{\mathrm{res}} \nonumber \\[1ex]
s^{n}_{t+1} &= \sigma\Big( \sum_{i=pre(n)}
    w_i \cdot s^{i}_{t} 
    +\sum_{j=post(n)} w_{j}^\top \cdot s^{j}_{t} 
\Big), \\
&\quad \text{for } N^{\mathrm{res}} < n < N \nonumber
\end{align}
where $N^{res}$ is the number of neuronal states in the Hopfield-Resnet blocks, and $N$ is the total number of neuronal states. The terms $pre(n)$ and $post(n)$ denote all previous and subsequent states relative to state $n$ that interact directly with it. This modification increases the number of trainable parameters and causes the network to take a longer time to reach steady state. Neuronal states interact through the weight parameters interconnecting them. Since the network is not feedforward, symmetric weight matrix is used in both directions during energy computation. Additionally, the newly computed neuronal pre-activation state is passed through an activation function to compute the updated neuronal state. Details of the activation function are provided in the following section.

\subsection{Alternative Activation Function}
Previous works relied on specialized activation functions; for example, in HEP~\cite{HEP}, the activation function is required to be holomorphic. Similarly, other implementations employed modified versions of the $sigmoid$ or $tanh$ functions, adjusted so that their outputs remain within the interval $[0,1]$. Bounding the energy function is essential to prevent explosive growth in energy values, which would otherwise destabilize the training process. We experimentally observed that existing activation functions limit the achievable accuracy as the network scales deeper. We propose a simplified bounded activation function, denoted as $\mathrm{ReLU}\alpha$, which restricts the output values to the range $[0, \alpha]$. A special case, $\mathrm{ReLU}1$, was previously introduced as an approximation to the $softmax$ function and shown to improve network accuracy~\cite{EnergyCompare}. In our experiments, we evaluated different initialization strategies for $\alpha$, including random initialization. We adopted $\mathrm{ReLU}6$ for the final experiments reported in this work.

\section{Results}
We validated the Hopfield-Resnet architecture on CIFAR-10, CIFAR-100, and Fashion-MNIST datasets. Centered equilibrium propagation (CEP, described in Section~\ref{sec:CEP}) was implemented using Nesterov accelerated gradient optimizer~\cite{nag}. %
The value of nudge parameter $\beta$ was tuned empirically. Larger values $(\beta\ge0.8)$ prevented learning progress, while smaller values (for example, $1e-4$) allowed training but introduced instabilities and increased sensitivity to other hyperparameters. In contrast, values in the range $[0.1,0.4]$ demonstrated greater stability during training. The experiments were performed on NVIDIA RTX 4090 and 6000 Ada GPUs using Pytorch.

As highlighted in Table~\ref{tab:model_dataset_comparison}, our experiments outperformed the state-of-the-art accuracy reported for the deep convolutional Hopfield network (DCHN)~\cite{EnergyCompare} trained using EP on both CIFAR10 and CIFAR100. The experimental findings further indicate that our approach yielded results much closer to those achieved by standard backpropagation (BP) training on similar network architectures. As shown in Table~\ref{tab:BPvsEP}, the performance gap between the two methods has been significantly narrowed, at times matching the performance of BP, demonstrating the viability of this alternative training paradigm.

\begin{table}[h!]
    \centering
    \begin{tabular}{|c|c|c|c|}
        \hline
        Dataset&Model Architecture&Prior Best (\%)&Our work (\%)\\
        \hline
        CIFAR-10       & VGG5       & 90.3 & 92.84 \\
               & Hopfield-Resnet13       & --  & 93.92\\
        CIFAR-100      & VGG5    & 68.4 & 70.78 \\
               & Hopfield-Resnet13     & --  & 71.05 \\
        F-MNIST   & VGG5   & 93.53 & 94.34 \\
           & Hopfield-Resnet13   & -- & 94.15 \\
        \hline
    \end{tabular}
    \caption{Accuracy comparison across models and datasets for the proposed architectural enhancements over the baseline implementation~\cite{EnergyCompare}.}
    \label{tab:model_dataset_comparison}
\end{table}

\begin{table}[h!]
    \centering
    \begin{tabular}{|c|c|c|c|}
        \hline
        Dataset & Model Architecture & BackProp (\%) & EquiProp (\%) \\
        \hline
        CIFAR-10       & VGG5       & 92.11 & 92.84 \\
               & Hopfield-Resnet13       & 93.78  & 93.92\\
        CIFAR-100      & VGG5    & 72.54 & 70.78 \\
               & Hopfield-Resnet13       & 75.12  & 71.05 \\
        \hline
    \end{tabular}
    \caption{Accuracy comparison of models trained with equilibrium propagation versus backpropagation. Note that BP was applied to the feedforward equivalent of the network.}
    
    \label{tab:BPvsEP}
\end{table}

\subsection{Ablation Studies}
\begin{figure}
    \centering
    \includegraphics[width=0.99\linewidth]{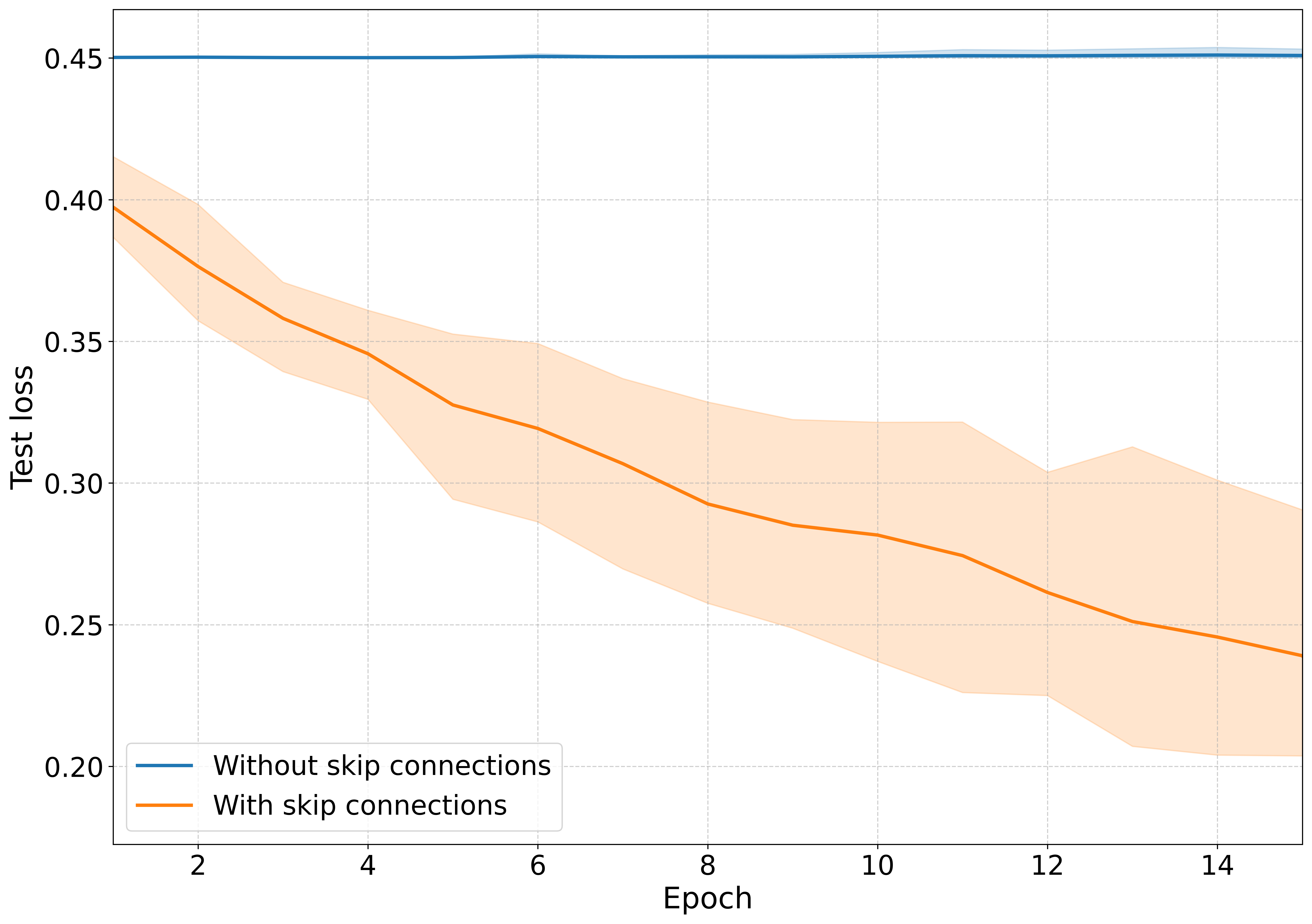}
    \caption{Test loss for the Hopfield-Resnet13 architecture trained using centered equilibrium propagation (CEP), with and without the skip connections, on the CIFAR-10 test set.}
    \label{fig:with and without skip}
\end{figure}

Our experiments showed that CEP struggles to perform well on deeper networks without the proposed skip connections. Having more than five consecutive convolutional layers slows down the training process without significant improvements in accuracy. Fig.~\ref{fig:with and without skip} shows the training loss progression for the aforementioned network architecture, both with and without skip connections, averaged over five experimental runs with varying hyperparameter combinations. Without skip connections, the training loss remains stagnant, showing negligible or no improvement over time. In contrast, when skip connections are included, the CEP algorithm successfully trains the model across a wide range of hyperparameters.

\begin{figure}
    \centering
    \includegraphics[width=0.99\linewidth]{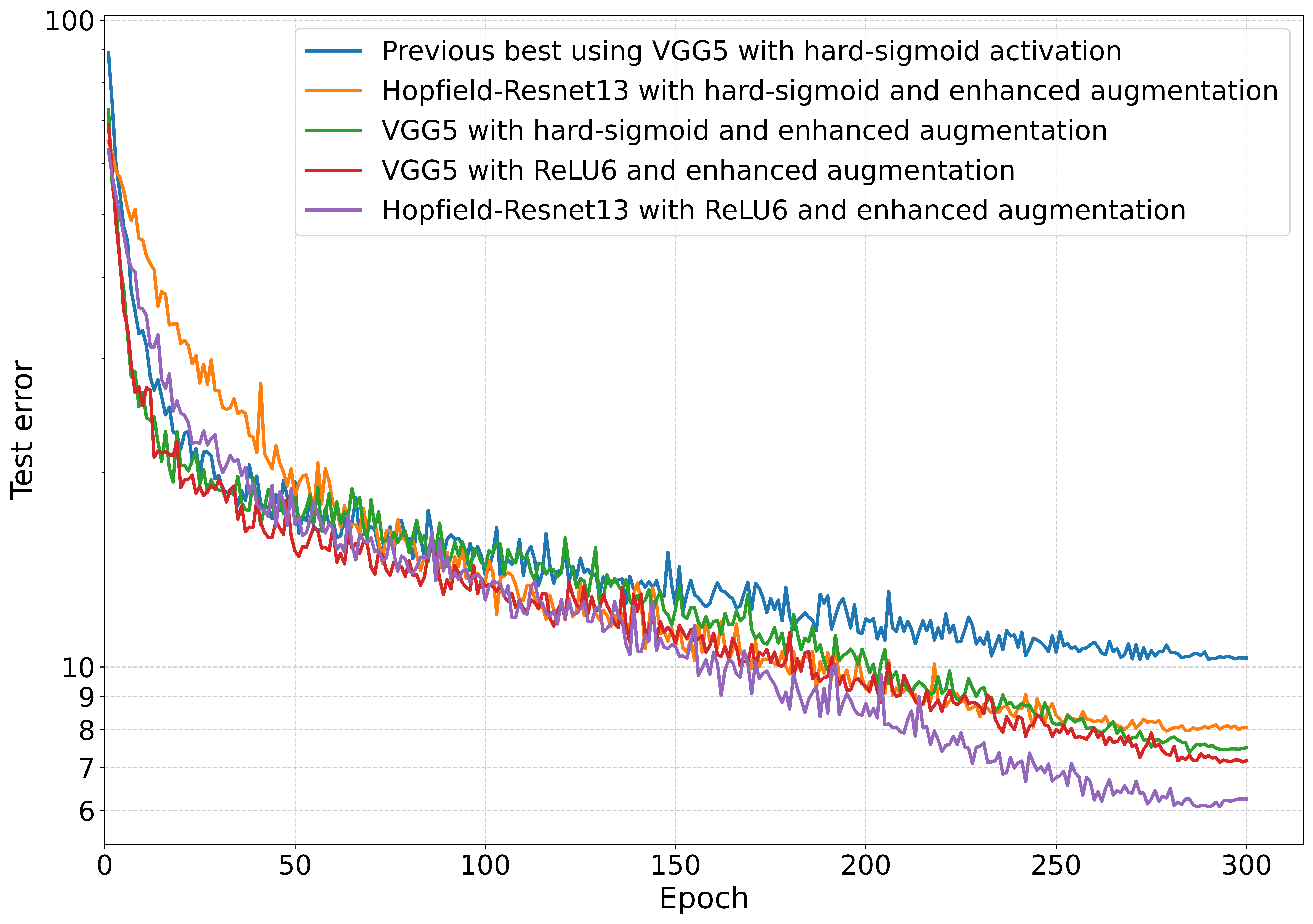}
    \caption{Test error over epochs for VGG5 and Hopfield-Resnet architectures, shown for different combinations of activation function and data augmentation on the CIFAR-10 test set. Note that the $y$-axis is displayed in $log$ scale.}
    \label{fig:compare}
\end{figure}

Regarding the activation function, we observed that setting $\alpha = 6$ ($\mathrm{ReLU}6$) yielded higher accuracy compared to $\mathrm{ReLU}1$ (hard-sigmoid function) for both VGG5 and Hopfield-Resnet architectures, as shown in Fig.~\ref{fig:compare}. The performance of $\mathrm{ReLU}{\alpha}$, where $\alpha$ is initialized uniformly at random to lie in the range $[0,10]$, was found to be between that of $\mathrm{ReLU}6$ and $\mathrm{ReLU}1$. Fig.~\ref{fig:compare} further demonstrates that the overall performance improvement results from the combination of all the suggested modifications. While enhancements to the data augmentation pipeline alone allowed our algorithm to surpass the previous best results on CIFAR-10, they did not improve the accuracy for larger networks. To achieve performance comparable to similarly sized networks trained with backpropagation, adjustments to the activation function were also necessary. These modifications, along with residual connections, enabled deeper networks to outperform shallower ones. In summary, the proposed Hopfield-Resnet architecture with residual connections, trained using centered equilibrium propagation with $\mathrm{ReLU}6$ activation and enhanced input data augmentation, achieved the best accuracy compared to similarly sized baseline model.

\subsection{Training Time}
We adopted the same training methodology as in previous works, using a fixed number of timesteps rather than running the simulation until true equilibrium was reached~\cite{EnergyCompare, ScalingToConv,HEP}. For both the best-performing VGG5 and Hopfield-ResNet13 architectures, we used $120$ timesteps in the free phase, followed by $50$ timesteps each with $+\beta$ and $-\beta$ in the weakly clamped phase. Increasing the number of timesteps to $200$ in the free phase and $60$ in the clamped phase, without incorporating data augmentation or activation changes, did not enable the network to achieve its best performance.

The training time for the best performing Hopfield-Resnet13 architecture, with $\mathrm{ReLU}6$ and enhanced data augmentation, exceeded $30$ hours for $300$ epochs on an RTX 6000 Ada GPU. Analysis of the program execution showed that a considerable amount of time was expended on kernel launches and CPU-GPU synchronization during GPU training. This indicates that optimization opportunities exist similar to those in feedforward networks, and that substantial reductions in training time can be achieved by optimizing the implementation of CEP training algorithm for GPU execution.

\subsection{Memory Utilization}
The memory utilization of centered equilibrium propagation (CEP) during training phase is comparable to that of a similar feedforward network trained using backpropagation (BP). For Hopfield-Resnet13 model, with a batch size of 128 on CIFAR-10, CEP required 1612 MiB of GPU memory on an RTX 4090. In contrast, the feedforward BP counterpart (Resnet13), which includes additional batch normalization layers, consumed 1324 MiB. Notably, Resnet13 without batch normalization significantly underperforms, achieving only $89\%$ test accuracy on the CIFAR-10 dataset. This suggests that equilibrium propagation, even without batch normalization, effectively handles data distribution through its energy function, thereby reducing the necessity for additional regularization.

\subsection{Weight Distribution}
\begin{figure*}
    \centering
    \includegraphics[width=1\linewidth]{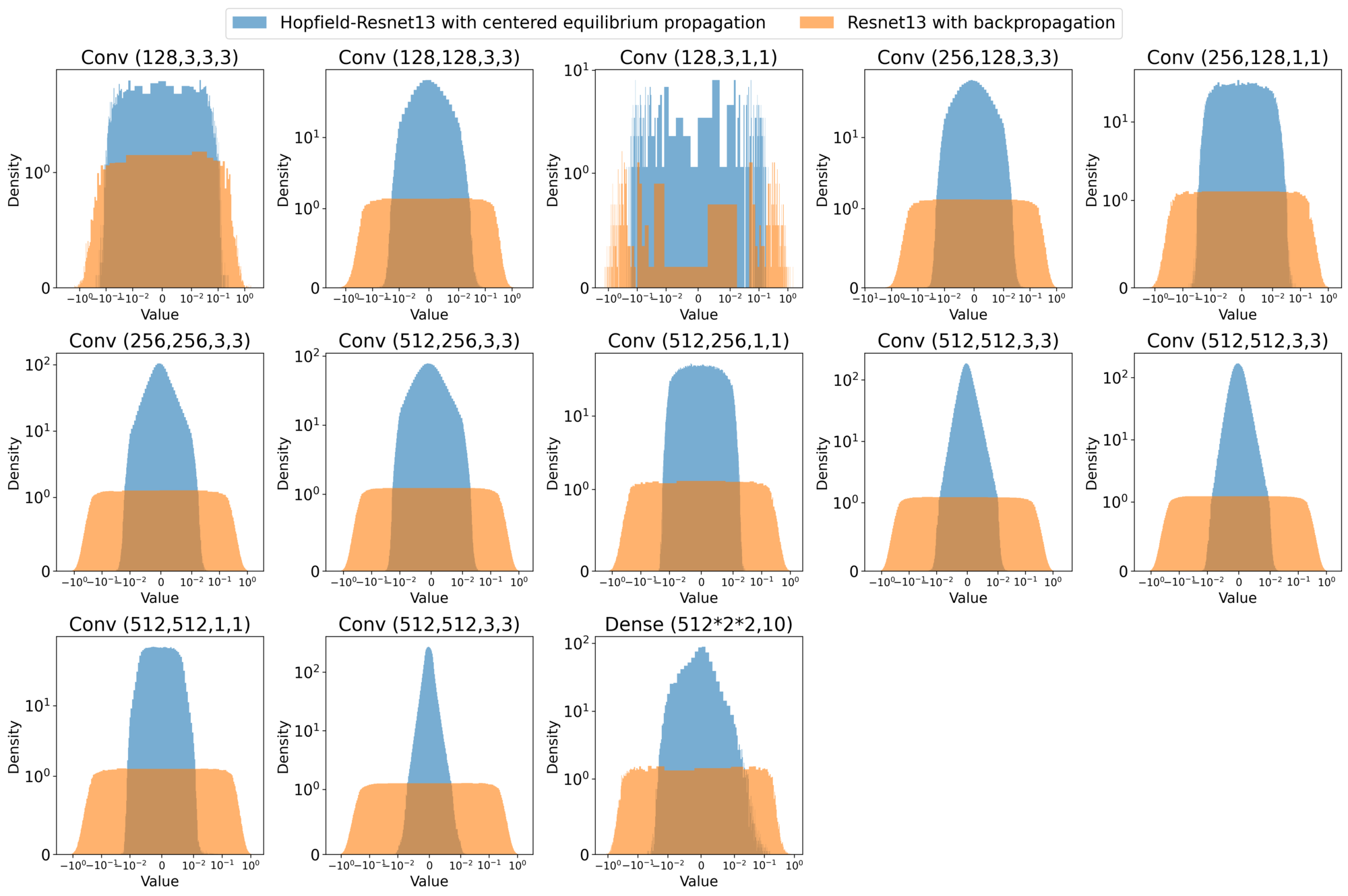}
    \caption{Layer wise distribution of weight values in  Resnet13 trained with backpropagation and Hopfield-Resnet13 trained with centered equilibrium propagation on the CIFAR-10 dataset.}
    \label{fig:weight}
\end{figure*}

Our experiments reveal that networks trained with centered equilibrium propagation (or CEP) exhibit distinctly different weight distributions compared to those obtained using backpropagation (or BP) training, underscoring their fundamentally different optimization objectives. CEP-trained networks tend to have both smaller absolute weight values and lower weight variance than their BP-trained counterparts. Fig.~\ref{fig:weight} illustrates these disparities: BP-trained networks exhibit a wider spread of weight values with relatively consistent distributions across layers, whereas CEP-trained networks have weights confined to a much narrower range. A notable trend emerges as network depth increases: in CEP-trained networks, weight magnitudes in deeper layers progressively approach zero, with this effect becoming more pronounced as depth increases. Although the initial layers maintain weight distributions similar to those in BP-trained networks, the deeper convolutional layers contain a much higher proportion of near-zero weights. The concentration of weights around zero can be partially attributed to weight decay, which is crucial for CEP performance. However, under the same weight decay setting, BP is unable to achieve the performance level of CEP. This tendency towards near-zero weight distributions could explain the difficulties CEP-trained models face when scaling to deeper architectures. Notably, the prevalence of near-zero values is considerably lower in layers with skip connections than in those along the direct connection path. This reduction in sparsity helps mitigate the challenges associated with training deeper networks using CEP.

\section{Conclusion}

In this work, we introduced residual connections between the hidden layers of convolutional Hopfield networks, enabling equilibrium propagation (EP) to scale to deeper networks than previously reported. In addition, we used $\mathrm{ReLU}{\alpha}$ as the non-linear activation instead of hard-sigmoid to improve training stability. These architectural enhancements yielded significant accuracy gains on the CIFAR-10, CIFAR-100, and Fashion MNIST datasets, surpassing previous results and approaching the performance of comparably sized feedforward networks trained with backpropagation (BP). Recent efforts to apply EP to modern Hopfield networks~\cite{hopfieldisallyouneed} for sequence learning~\cite{SeqEP} highlight exciting directions for future research. Integrating these approaches with the architectures proposed in this work, and investigating other unique properties resulting from EP’s distinct training methodology, represent promising avenues to extend EP’s applicability.

Despite these promising directions, the practical adoption of EP as an alternative to BP faces notable challenges. Training deeper networks with EP remains computationally intensive due to the inherently sequential nature of the algorithm and the limitations of current GPU architectures, which are optimized for parallel processing rather than iterative computations. To fully realize the potential of EP, two critical advancements are required: the development of specialized hardware tailored to EP’s computational demands, and algorithmic optimizations that better leverage existing hardware capabilities. Such innovations could significantly reduce training times and establish EP as a viable and efficient alternative to BP.

\bibliographystyle{IEEEtran}
\bibliography{references}

\end{document}